\documentclass[a4paper,10pt,twocolumn]{ieeeconf}                                                         

\IEEEoverridecommandlockouts                              

\usepackage{graphicx}          

\usepackage{times,mathptmx}
\usepackage[T1]{fontenc}
\usepackage{color}
\usepackage{epsfig}
\usepackage[latin1]{inputenc}
\usepackage{rotating}
\usepackage{amsmath}
\usepackage{amssymb}
\usepackage{amsfonts}
\usepackage{epsf}
\usepackage{graphicx}
\usepackage{psfrag}
\usepackage{subfigure}
\usepackage{url}
\usepackage{amsbsy}

\title{\LARGE \bf On $l_1$ Mean and Variance Filtering}

\author{Bo Wahlberg, Cristian R. Rojas  and Mariette Annergren\\
Automatic Control Lab and ACCESS, School of Electrical Engineering,\\ KTH Royal Institute of Technology, SE-100 44 Stockholm, Sweden.\\
 \thanks{\mbox{e-mails:~bo.wahlberg@ee.kth.se,~cristian.rojas@ee.kth.se,} \mbox{$\qquad\qquad\;$ mariette.annergren@ee.kth.se}.}
\thanks{This work was partially supported by the Swedish Research Council and the Linnaeus Center ACCESS at KTH. The research leading to these results has received funding from The European Research Council under the European Community's  Seventh Framework program (FP7 2007-2013) / ERC Grant Agrement N. 267381}
}

\begin{document}

\maketitle \thispagestyle{empty} \pagestyle{empty}

\begin{abstract}
This paper addresses the problem of segmenting a time-series with respect to changes in the mean value or in the variance. The first case is when the time data is modeled as a sequence of independent and normal distributed random variables with unknown, possibly changing, mean value but fixed variance. The main assumption is that the mean value is piecewise constant in time, and the task is to estimate the change times and the mean values within the segments. The second case is when the mean value is constant, but the variance can change. The assumption is that the variance is piecewise constant in time, and we want to estimate change times and the variance values within the segments. To find solutions to these problems, we will study an $l_1$ regularized maximum likelihood method, related
to the fused lasso method and $l_1$  trend filtering, where the parameters to be estimated are free to vary at each sample. To penalize variations in the estimated parameters, the
$l_1$-norm of the time difference of the parameters is used as a regularization term. This idea is closely related to total variation denoising. The main contribution is that
 a convex formulation of this variance estimation problem, where the parametrization is based on the
inverse of the variance, can be formulated as a certain $l_1$ mean estimation problem. This implies that results and  methods for mean estimation can be applied to the challenging problem of variance segmentation/estimation.

\end{abstract}
{\em Copyright 1998 IEEE. Published in the Proceedings of the 45th Annual Asilomar Conference on Signals, Systems, and Computers,  November 6-9, 2011, Pacific Grove, California, USA
Personal use of this material is permitted. However, permission to reprint/republish this material for advertising or promotional purposes or for creating new collective works for resale or redistribution to servers or lists, or to reuse any copyrighted component of this work in other works, must be obtained from the IEEE.}


\section {Introduction}
The problem of estimating the mean, trends and variances in time
series data is of fundamental importance in signal processing and
in many other disciplines such as processing of financial and
biological data. This is typically done to preprocess data before
estimating for example parametric models. For non-stationary data
it is also important to be able to detect changes in mean and
variances and segment the data into stationary subsets. A
classical way is to use windowing to handle time-variations, by
for example subtracting the windowed sample mean estimate from the
data or scaling the data with a windowed estimate of the variance.
More advanced detection and segmentation methods are often based
on probabilistic models, such as Markov models, and have a lot of
tuning and user choices, \cite{Gustaffson-01}. An alternative way
to approach this problem is to use regularization to penalize
variations and changes in the estimated parameter vector.
Recently, there has been a lot of efforts on applying $l_1$-norm
regularization in estimation in order to obtain convex optimization problems, \cite{citeulike:161814,DBLP:books/daglib/0025129}.
Our work is
inspired by the $l_1$ trend filtering method in
\cite{Kim-Koh-Boyd-Gorinevsky-09} and the {fused lasso}
method, \cite{Tibshirani-Saunders-Rosset-Zhu-Knight-05}. The $l_1$
trend filtering method considers changes in the mean value of the
data. Here we are also interested in changes in the variance. This
problem is closely related to the covariance selection problem
introduced in \cite{Dempster-72}. The paper
\cite{Banerjee-ElGhaoui-dAspremont-08} formulates this as a convex
optimization problem by using the inverse of the covariance matrix
as parameter, see also \cite{Kim-Koh-Boyd-Gorinevsky-09}. This
idea is also used in the { graphical lasso method},
\cite{Friedman-Hastie-Tibshirani-08}.

The paper is organized as follows. In  Section \ref{sec:1} the general problem formulation is specified. Section  \ref{sec:2} considers the special case of mean estimation, while
 Section  \ref{sec:3} deals with variance estimation and its relation to mean estimation.  Section   \ref{sec:4} contains a numerical example of variance estimation, and Section  \ref{sec:5} discusses the extension to the multi-variate case. The paper is concluded  in Section \ref{sec:6}.
\section{Problem Statement}
\label{sec:1}
Consider the independent scalar sequence $\{y_t\}$ which satisfies $$y_t \sim
\mathcal{N}(m_t, \sigma^2_t),$$
where both the mean $\{m_t\}$ and the variance $\{\sigma^2_t\}$ are (unknown) piecewise constant sequences.
Assume that the measurements $\{y_1 \; \cdots \; y_N\}$
are available, and we are interested in estimating $m_1, \ldots,
m_N$ and $\sigma^2_1, \ldots, \sigma^2_N$.

To solve this problem, first notice that the model $y_t \sim
\mathcal{N}(m_t, \sigma_t)$ is a \emph{standard exponential family
with canonical parameters}, \cite[Example~1.2]{Brown-86},
 $$\mu_t := m_t / \sigma_t^2 \in
\mathbb{R},\quad \eta_t := - 1 / 2\sigma_t^2 \in \mathbb{R}^-$$
This means that the log-likelihood
of $\{\mu_1, \dots, \mu_N,$ $ \eta_1, \dots, \eta_N\} $ given
$\{y_1 \; \cdots \; y_N\}$ is
\begin{align*}
{l(\mu_1, \dots, \mu_N, \eta_1, \dots, \eta_N)=}& - \frac{N}{2} \ln \pi  + \sum_{t = 1}^N \left[\frac{\ln(-\eta_t)}{2}\right.\\ &  \left.+ \frac{\mu_t^2}{4 \eta_t} + \eta_t y_t^2 + \mu_t y_t\right]. %
\end{align*}
Moreover, by \cite[Theorem 1.13]{Brown-86} it follows that $l$ is strictly
concave on $$\{(\mu_1, \dots, \mu_N, \eta_1, \dots, \eta_N):\;
\mu_t \in \mathbb{R},\; \eta_t \in \mathbb{R}^-,\; t = 1, \dots
,N\}.$$

{\bf Assumption:} The prior knowledge  is that the sequences $\{m_t\}$ and $\{\sigma^2_t\}$ are piecewise constant in time. This means that the difference sequences
$\{m_{t+1}-m_t\}$ and $\{\sigma^2_{t+1}-\sigma^2_t\}$ are sparse.

Inspired by \cite{Kim-Koh-Boyd-Gorinevsky-09} we propose an estimator based
on the solution of the following optimization problem:
\begin{align} \label{eq:1}
\underset{\begin{array}{c} \scriptstyle \mu_1, \dots, \mu_N \\ \scriptstyle \eta_1, \ldots, \eta_N \end{array}}{\text{minimize}} & \left\{
 \sum_{t = 1}^N \left[\frac{-\ln(-\eta_t)}{2} -  \frac{\displaystyle \mu_t^2}{\displaystyle 4 \eta_t}  -  \eta_t y_t^2 - \mu_t y_t\right]\right.\nonumber \\ & \left. + \lambda_1 \sum_{t = 2}^N |\mu_t - \mu_{t - 1}| + \lambda_2 \sum_{t = 2}^N |\eta_t - \eta_{t - 1}|\right\}\\ %
\text{subject to } & %
\eta_t  < 0, \quad t = 1, \dots, N. \nonumber%
\end{align}
This a convex optimization problem where the cost function is separable, plus two terms that are separable
in the difference between consecutive variables. The $l_1$-norm is used to penalize non-sparse solutions, while still having a convex objective function. Standard software for convex optimization can be used to solve (\ref{eq:1}). However, it is possible to use the special structure to derive more efficient optimization algorithms for (\ref{eq:1}), see \cite{DBLP:journals/ftml/BoydPCPE11,BW_SYSID12}.

\section{Mean Estimation}
\label{sec:2}
Consider the problem of mean segmentation under the assumption that the variance is constant. The optimization problem (\ref{eq:1}) then simplifies to
\begin{align} \label{eq:12}
\underset{\begin{array}{c} \scriptstyle m_1, \dots, m_N  \end{array}}{\text{minimize}}\; & V(m_1,\ldots, m_N),
\end{align} where
\begin{align}
V(m_1,\ldots, m_N)&=
 \frac{1}{2}\sum_{t = 1}^N (y_t-m_t)^2+  \lambda_1 \sum_{t = 2}^N |m_t - m_{t - 1}| .
\end{align}
The $t$:th element of the sub-differential of the cost function equals
\begin{align}
[{\partial}V(m_1,\ldots, m_N)]_t&  = y_t-m_t +\lambda_1 [{\rm sign}(m_t - m_{t - 1})\nonumber\\ & -{\rm sign}(m_{t+1} - m_{t})],\; 2<t<N-1,
\label{eq:sg}
\end{align}
where
\begin{equation}
 {\rm sign}(x)\left\{\begin{array}{lr}= -1, & x<0,\\ \in {[}-1,1{]}, &  x=0,\\ =1, & x>0.
\end{array}\right.
\label{eq:sign}
\end{equation}
For $t=1$ and $t=N$ obvious modifications have to be done to take the initial and end conditions into account. Using the incremental structure of the sub-differential, it makes sense to add up the expressions(\ref{eq:sg}) to obtain
\begin{align*}
\sum_{t=1}^k[{\partial}V(m_1,\ldots, m_N)]_t&=\sum_{t=1}^k(y_t-m_t)-\lambda_1 {\rm sign}(m_{k+1} - m_{k}),\\ & \qquad \qquad\; 1\leq k<N,\\
\sum_{t=1}^N[{\partial}V(m_1,\ldots, m_N)]_t&=\sum_{t=1}^N(y_t-m_t).
\end{align*}
This is more or less the sub-differential with respect to the variables $r_1= m_1$, $r_{k}=m_{k}-m_{k-1}$, $k=2,\ldots, N $.
For optimality the sub-gradient should include zero,  which leads to the optimality conditions
\begin{align}
\sum_{t=1}^k(y_t-m_t)&=\lambda_1 {\rm sign}(m_{k+1} - m_{k}), \quad 1\leq k<N,\label{eq:op}\\
\sum_{t=1}^N(y_t-m_t)&=0.\label{eq:mean}
\end{align}
The "empirical mean"
\begin{equation}
\widehat{m}=\frac{1}{N}\sum_{t = 1}^N y_t
\end{equation}
obtained from (\ref{eq:mean}),
satisfies also the first $N-1$ optimality conditions (\ref{eq:op}) if
\begin{align*}
\left|\sum_{t=1}^k(y_t-\widehat{m})\right|&=\left|\sum_{t=1}^ky_t-\frac{k}{N}\sum_{t = 1}^N y_t\right|\leq \lambda_1, \; 1\leq k<N.
\end{align*}
Here we have used that  ${\rm sign}(0)\in  {[}-1,1{]}$. This is the case if $\lambda_1$ is large enough. Since the optimization problem \eqref{eq:12} is convex, the sub-differential condition is necessary and sufficient. Hence we have now derived the $\lambda_{\max}$ result, \cite{Kim-Koh-Boyd-Gorinevsky-09}
\begin{align}
[\lambda_1]_{\max} = \max_{k = 1, \ldots, N - 1} \left| \sum_{t = 1}^k y_t - \frac{k}{N} \sum_{t = 1}^N y_t \right|. \label{eq_maxx}%
\end{align}
Then, $m_1 = \cdots = m_N = \widehat{m}$ is the optimal
solution of \eqref{eq:12} if and only if $\lambda_1 \geq [\lambda_1]_{\max}$.
The expression for $[\lambda_1]_{\max}$ is more obvious by diving (\ref{eq_maxx}) by $k$,
$$
\frac{1}{k} \sum_{t = 1}^k y_t - \frac{1}{N} \sum_{t = 1}^N y_t.
$$
Hence, we compare the empirical means for the sequences of length $k = 1, \ldots, N - 1 $ with $\widehat{m}$, and then relate $\lambda$ to the maximum deviation.

The  $\lambda_{\max}$ result is very useful in order to find a good choice of $\lambda$, and also to derive efficient numerical solvers.
\section{Variance Estimation}
\label{sec:3}
We will now study the variance estimation problem under the assumption that the mean values are known. We can,  without losing generality, assume that $m_t=0$.
For this special case the optimization problem (\ref{eq:1}) equals
\begin{align} \label{eq:13}
\underset{\begin{array}{c} \scriptstyle \eta_1, \dots, \eta_N  \end{array}}{\text{minimize}}\; & W(\eta_1,\ldots, \eta_N),\\
\text{subject to } & %
\eta_t  < 0, \quad t = 1, \dots, N,\nonumber %
\end{align}
where
\begin{align}
W(\eta_1,\ldots, \eta_N)&=  \sum_{t = 1}^N \left[ \frac{-\ln(-\eta_t)}{2} -\eta_t y_t^2\right]  + \lambda_2 \sum_{t = 2}^N |\eta_t - \eta_{t - 1}|\nonumber\\
\end{align}

We will now show that (\ref{eq:13}) is equivalent, in the sense of having related optimal solution, to the optimization problem
\begin{align} \label{eq:14}
\underset{\begin{array}{c} \scriptstyle \sigma^2_1, \dots, \sigma^2_N  \end{array}}{\text{minimize}}\; & V(\sigma^2_1,\ldots, \sigma^2_N),\\
\text{subject to } & %
\sigma^2_t  >0, \quad t = 1, \dots, N,\nonumber %
\end{align}
where
\begin{align}
V(\sigma^2_1,\ldots, \sigma^2_N)&=
 \frac{1}{2}\sum_{t = 1}^N (y^2_t-\sigma^2_t)^2+  \lambda_2 \sum_{t = 2}^N |\sigma^2_t - \sigma^2_{t - 1}| .
\end{align}
and $\sigma_t^2=-1/(2\eta_t)$ is the variance. Now
\begin{equation}
\eta_t - \eta_{t - 1}=\frac{\sigma^2_t-\sigma^2_{t-1}}{2\sigma^2_t\sigma^2_{t-1}}\label{eq:sign},
\end{equation}
which means that the sign of the differences is not affected by the transformation  $\sigma_t^2=-1/(2\eta_t)$. This will be critical in deriving the equivalence result.
The formulation  (\ref{eq:14}) also makes sense from a practical point of view, since the variance of $y_t$ (in the zero mean case) is the mean of $y^2_t$. Notice, however, that this is not directly obvious from the log-likelihood, but is often used in signal processing under the name of covariance fitting, \cite{Stoica-Moses-05}.

We now have the following main result:

\vspace{5pt}
\textbf{Theorem 1.} \emph{The convex optimization problems \eqref{eq:13} and \eqref{eq:14}, with $\sigma_t^2=-1/(2\eta_t)$, have the same sub-gradient optimality conditions.}
\vskip\baselineskip\noindent
{\bf Proof:} First notice that
\begin{align*}
\left[ {\partial}\sum_{t = 1}^N \left(\frac{-1}{2}\ln(-\eta_t) -\eta_t y_t^2\right) \right]_t&=\frac{-1}{2\eta_t}-y^2_t\\
=\sigma_t^2-y_t^2&= \left[{\partial}\frac{1}{2} \sum_{t = 1}^N (\sigma_t^2 -y_t^2)^2 \right]_t.
\end{align*}
Next, for $2\leq t<N$
\begin{align*}
&\left[{\partial}\sum_{t = 2}^N |\eta_t - \eta_{t - 1}|\right]_t={\rm sign}(m_t - m_{t - 1}) -{\rm sign}(m_{t+1} - m_{t})\\&=
{\rm sign}(\sigma^2_t - \sigma^2_{t - 1}) -{\rm sign}(\sigma^2_{t+1} - \sigma^2_{t})=\left[{\partial}\sum_{t = 2}^N |\sigma^2_t - \sigma^2_{t - 1}|\right]_t.
\end{align*}
Here we have used that the sign function defined by (\ref{eq:sign}) only depends on the sign of its argument and (\ref{eq:sign}) implies that the sign is not changed by the transformation $\sigma_t^2=-1/(2\eta_t)$.\\ $\makebox[1cm]{}$\hfill {\bf Q.E.D.}

Since both optimization problems \eqref{eq:13} and \eqref{eq:14} are convex, Theorem 1 implies  that we can re-use algorithms and results for the mean estimation problem to the variance estimation problem \eqref{eq:14}.
For example, it directly follows that
\begin{align*}
[\lambda_2]_{\max} = \max_{k = 1, \ldots, N - 1} \left| \sum_{t = 1}^k y^2_k - \frac{k}{N} \sum_{t = 1}^N y^2_t \right|,%
\end{align*}
and for $\lambda_2\geq [\lambda_2]_{\max}$  the constant  "empirical variance" solution
\begin{equation}
\widehat{\sigma}^2=\frac{1}{N}\sum_{t = 1}^N y^2_t,\quad \widehat{\eta}=\frac{-1}{2\widehat{\sigma}^2}
\end{equation}
are the optimal solutions to \eqref{eq:13} and \eqref{eq:14}, respectively. From a practical point of view one has to be a bit careful when squaring $y_t$ since outliers are amplified.

\section{Example}
\label{sec:4}
\begin{figure}[ht]
\begin{center}
\includegraphics[width = 0.9\columnwidth]{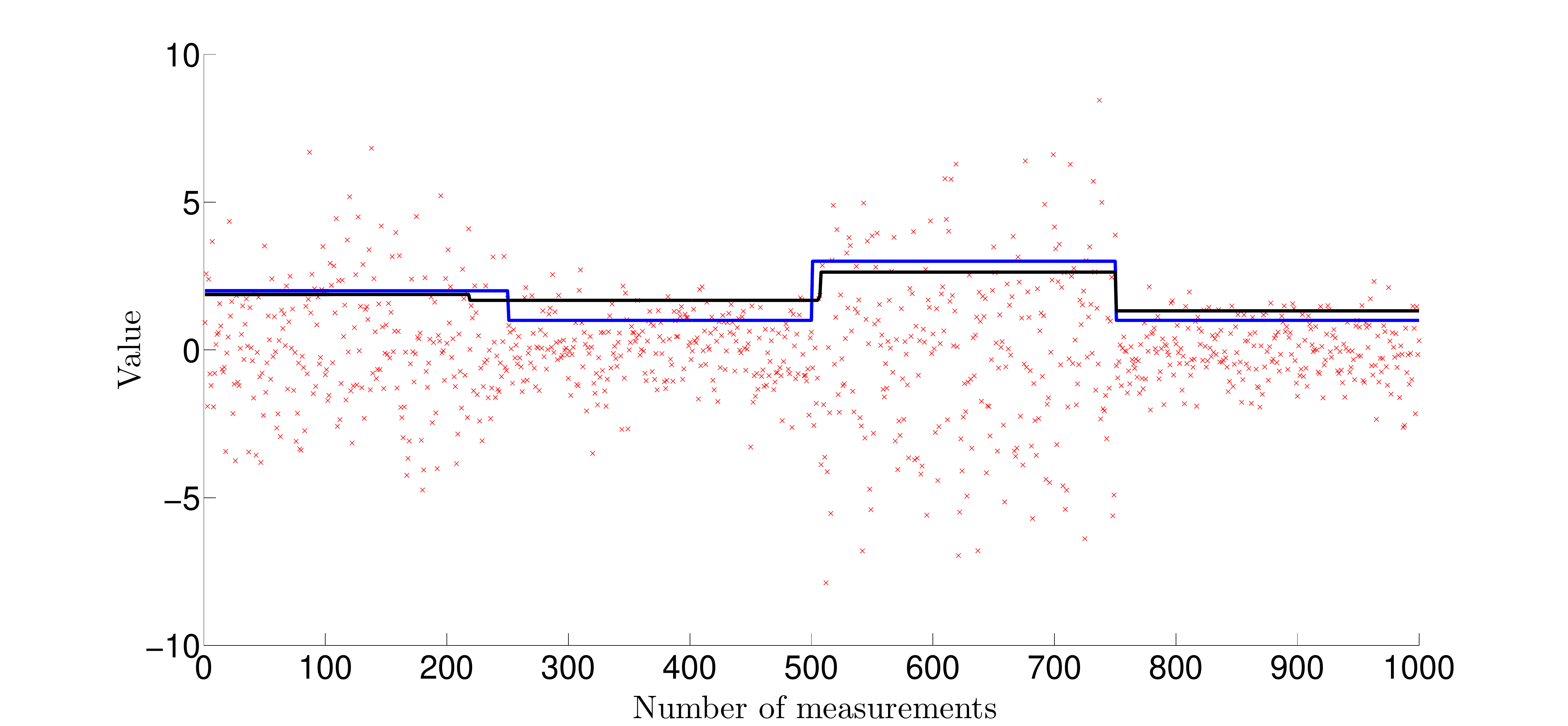}
\caption{Estimated variance (black line), true variance (blue
line) and measurements (red crosses).} %
\label{ex1} %
\end{center}
\end{figure}

Consider a signal $\{y_t\}$ which satisfies $y_t\sim
\mathcal{N}(0,\sigma^2_t)$, where $\{\sigma^2_t\}$ is a piece-wise
constant sequence:
\begin{align*}
\sigma^2_t = \left\{
\begin{array}{rl}
2, & \text{if } 0< t \leq 250,\\
1, & \text{if } 250 < t \leq 500,\\
3, & \text{if } 500 < t \leq 750,\\
1, & \text{if } 750 < t \leq 1000.
\end{array} \right.
\end{align*}
Given $1000$ measurements $\{y_1, \cdots , y_{1000}\}$, we want to estimate the variances $\sigma^2_1, \ldots,
\sigma^2_{1000}$. To solve problem~\eqref{eq:14} we used
\texttt{CVX}, a package for specifying and solving convex programs
\cite{Grant-Boyd-08}.
Figure~\ref{ex1} shows the resulting
estimates of  $\{\sigma^2_t\}$, the true values of $\{\sigma^2_t\}$
and the measurements $\{y_1, \cdots , y_{1000}\}$.

\section{Two Extensions}
\label{sec:5}
\subsection{Simultaneous Mean and Variance Estimation}
The general mean and variance optimization problem (\ref{eq:1}) is convex and it is possible to find $\lambda_{\max}$ expressions. A difficulty is the term
${\displaystyle \mu_t^2}     /{\displaystyle 4 \eta_t}$ in (\ref{eq:1})
that couples the mean and variance optimization.  It is also non-trivial to tune this algorithm in the sense that it is difficult to separate a change in mean from a change in variance based on short data records.

\subsection{The Multi-Variate Case}

Assume that the process $\mathbf{y}_t\sim
\mathcal{N}(\mathbf{m}_t,\Sigma_t)\in \mathbb{R}^n$, that is the mean $\mathbf{m}_t\in\mathbb{R}^n$ and the covariance matrix  $\Sigma_t\in \mathbb{R}^{n \times n}$. The canonical parameters are
 $$\mathbf{\mu}_t := \Sigma_t^{-1}\mathbf{m}_t  \quad \mathbf{H}_t := \frac{-1}{2} \Sigma_t^{-1}.$$
 The  corresponding $l_1$ regularized maximum log-likelihood estimation problem is
\begin{align*}
 \underset{\begin{array}{c} \scriptstyle \mathbf{\mu}_1,\ldots, \mathbf{\mu}_N\\ \scriptstyle \mathbf{H}_1<0,\ldots\mathbf{H}_N< 0\end{array}}{\mbox{minimize}} & \: \left\{\sum_{i=1}^N \left( \frac{-1}{2}\log\det\{-\mathbf{H}_t\}-\frac{1}{4}\mbox{Tr}\{\mathbf{H}_t^{-1} \mathbf{\eta}_t\mathbf{\eta}_t^T\} \right.\right.\\& \left.\left.-\mbox{Tr}\{\mathbf{H}_t\mathbf{y}_y\mathbf{y}_t^T\} -\mathbf{\eta}_t^T \mathbf{y}_t \right. \right)\\ &\left. +\lambda_1 \sum_{i=2}^{N} \| \mathbf{\mu}_{t+1}-\mathbf{\mu}_t\|_2+\lambda_2 \sum_{i=2}^{N} \|\mathbf{H}_{t+1}-\mathbf{H}_t   \|_F\right\}
\end{align*}
where we have used the Euclidean vector norm and Frobenius matrix norm.  This is a convex optimization problem with a large number of unknowns, $n+(n+1)n/2$ per $n$ dimension sample $\mathbf{y}_t$.

A problem when trying to generalize the results on the equivalence of variance estimation and mean estimation of $\mathbf{y}_t\mathbf{y}_t^T$ is that the ordering relation
$$
\mathbf{H}_{t+1} -\mathbf{H}_t =\frac{1}{2}\Sigma_t^{-1}[\Sigma_{t+1}-\Sigma_t]\Sigma_{t+1}^{-1}
$$
does not holds componentwise. Still, the convex problem
\begin{align*}
 \underset{\begin{array}{c} \ \scriptstyle \Sigma_1>0,\ldots, \Sigma_n> 0\end{array}}{\mbox{minimize}} & \:\sum_{i=1}^N \|\mathbf{y}_t\mathbf{y}_t^T-\Sigma_{t}\|_F+\lambda_2 \sum_{i=2}^{N} \|\Sigma_{t+1}-\Sigma_t   \|_F
\end{align*}
makes sense as a covariance matrix fitting problem.

\section{Conclusions}
\label{sec:6}
The objective of this contribution has been to introduce the concept of $l_1$ variance filtering and relate this approach to the problem of $l_1$ mean filtering.
The advantage of the $l_1$ approach is that there are only one or two design parameters ($\lambda_1$ and $\lambda_2$), while classical approaches involve more user design variables such as thresholds and transition probabilities. The framework presented can also be used for more advanced time-series model estimations such as autoregressive  models, see \cite{OhlssonLB:10}.
The number of variables in the multi-variate problem can be huge. Tailored algorithms for this problem based on the alternating direction method of multipliers algorithm, \cite{DBLP:journals/ftml/BoydPCPE11}, have recently been proposed in~\cite{BW_SYSID12}.
\bibliography{refs_cristian,refs_bo}

\begin{thebibliography}{10}

\bibitem{Banerjee-ElGhaoui-dAspremont-08}
O.~Banerjee, L.~{El Ghaoui}, and A.~{d'Aspremont}.
\newblock Model selection through sparse maximum likelihood estimation for
  multivariate gaussian or binary data.
\newblock {\em Journal of Machine Learning Research}, 9:485--516, 2008.

\bibitem{DBLP:journals/ftml/BoydPCPE11}
S.~Boyd, N.~Parikh, E.~Chu, B.~Peleato, and J.~Eckstein.
\newblock Distributed optimization and statistical learning via the alternating
  direction method of multipliers.
\newblock {\em Foundations and Trends in Machine Learning}, 3(1):1--122, 2011.

\bibitem{Brown-86}
L.~D. Brown.
\newblock {\em Fundamentals of Statistical Exponential Families, With
  Applications in Statistical Decision Theory}.
\newblock Institute of Mathematical Statistics, 1986.

\bibitem{Dempster-72}
A.~P. Dempster.
\newblock Covariance selection.
\newblock {\em Biometrics}, 28(1):157--175, 1972.

\bibitem{DBLP:books/daglib/0025129}
M.~Elad.
\newblock {\em Sparse and Redundant Representations - From Theory to
  Applications in Signal and Image Processing}.
\newblock Springer, 2010.

\bibitem{Friedman-Hastie-Tibshirani-08}
J.~Friedman, T.~Hastie, and R.~Tibshirani.
\newblock Sparse inverse covariance estimation with the graphical lasso.
\newblock {\em Biostatistics}, 9(3):432--441, 2008.

\bibitem{Grant-Boyd-08}
M.~C. Grant and S.~P. Boyd.
\newblock Graph implementations for nonsmooth convex programs.
\newblock In V.~D. Blondel, S.~P. Boyd, and H.~Kimura, editors, {\em Recent
  Advances in Learning and Control (tribute to M. Vidyasagar)}, pages 95--110.
  Springer-Verlag, 2008.

\bibitem{Gustaffson-01}
F.~Gustafsson.
\newblock {\em Adaptive Filtering and Change Detection}.
\newblock Wiley, 2001.

\bibitem{citeulike:161814}
T.~Hastie, R.~Tibshirani, and J.~H. Friedman.
\newblock {\em {The Elements of Statistical Learning}}.
\newblock Springer, July 2003.

\bibitem{Kim-Koh-Boyd-Gorinevsky-09}
S.-J. Kim, K.~Koh, S.~Boyd, and D.~Gorinevsky.
\newblock $l_1$ trend filtering.
\newblock {\em SIAM Review}, 51(2):339--360, 2009.

\bibitem{OhlssonLB:10}
H.~Ohlsson, L.~Ljung, and S.~Boyd.
\newblock Segmentation of {ARX}-models using sum-of-norms regularization.
\newblock {\em Automatica}, 46:1107 -- 1111, April 2010.

\bibitem{Stoica-Moses-05}
P.~Stoica and R.~Moses.
\newblock {\em Spectral Analysis of Signals}.
\newblock Prentice Hall, Upper Saddle River, New Jersey, 2005.

\bibitem{Tibshirani-Saunders-Rosset-Zhu-Knight-05}
R.~Tibshirani, M.~Saunders, S.~Rosset, J.~Zhu, and K.~Knight.
\newblock Sparsity and smoothness via the fused lasso.
\newblock {\em Journal of the Royal Statistical Society: Series B (Statistical
  Methodology)}, 67 (Part 1):91--108, 2005.

\bibitem{BW_SYSID12}
B.~Wahlberg, S.~Boyd, M~Annergren, and W.~Yang.
\newblock An {ADMM} algoritm for a class of total variation regularized
  estimation problems.
\newblock {\em SYSID 2012}, 2011.
\newblock Submitted.

\end{thebibliography}

\end{document}